\pdfoutput=1

\documentclass[11pt]{article}

\usepackage[]{acl}

\usepackage{times}
\usepackage{latexsym}
\usepackage{booktabs}
\usepackage{multirow}
\usepackage{graphicx}
\usepackage{siunitx}
\usepackage{courier}
\usepackage{nameref}
\usepackage{caption}
\usepackage{subcaption}

\usepackage[T1]{fontenc}

\usepackage[utf8]{inputenc}

\usepackage{microtype}

\title{What is the best recipe for character-level encoder-only modelling?}

\author{Kris Cao \\
  DeepMind, London, UK \\
  \texttt{kriscao@deepmind.com}}

\begin{document}
\maketitle
\begin{abstract}
This paper aims to benchmark recent progress in language understanding models that output contextualised representations at the character level. Many such modelling architectures and methods to train those architectures have been proposed, but it is currently unclear what the relative contributions of the architecture vs. the pretraining objective are to final model performance. We explore the design space of such models, comparing architectural innovations \citep{Clark:22,jaegle2022perceiver,Tay:2021}, and a variety of different pretraining objectives on a suite of evaluation tasks in order to find the optimal way to build and train character-level BERT-like models. We find that the best recipe combines the Charformer and CANINE model architectures, and follows the CANINE training procedure. This model exceeds the performance of a token-based model trained with the same settings on the same data, suggesting that character-level models are ready for more widespread adoption. Unfortunately, the best method to train character-level models still relies on a learnt tokeniser during pretraining, and final model performance is highly dependent on tokeniser quality. We believe our results demonstrate the readiness of character-level models for multilingual language representation, and encourage NLP practitioners to try them for their needs.
\end{abstract}

\section{Introduction}
\begin{table*}[t]
    \centering
    \begin{tabular}{p{3cm} | p{3cm} | p{2.5cm} | p{2.5cm} | p{2.5cm}}
        \toprule
        \textbf{Input units} (\S \ref{sec:chars_or_bytes}) & \textbf{Downsampling model} (\S \ref{para:downsampling}) & \textbf{Upsampling model} (\S \ref{para:upsampling}) & \textbf{Prediction targets} (\S \ref{para:prediction_targets}) & \textbf{Masking scheme} (\S \ref{para:masking_schemes}) \\
        \midrule
        Characters with fixed embeddings & CANINE & CANINE & Tokens & Tokens \\
        \midrule
        Characters with learnt embeddings & Charformer & Perceiver & Independent characters & Whitespace \\
        \midrule
        Bytes & Perceiver & & Autoregressive characters & Random \\
        \bottomrule
    \end{tabular}
    \caption{An overview of all the design choices we examine for building character-level models. We compare the combinatorial space spanned by these building blocks in our experiments.}
    \label{tab:my_label}
\vspace{-1.5em}
\end{table*}
The first stage of almost all NLP modelling pipelines is to convert input text strings into a sequence of symbols that the model can ingest. This step, called tokenisation, can be highly non-trivial and introduces significant theoretical and practical complexities to both training and using these models. One particular issue for massively multilingual models is that many languages have to compete for space given a fixed vocabulary size, which limits the effective vocabulary size per language: as an illustration, the WordPiece tokeniser that multilingual BERT uses tokenises `hello' as two tokens: `hell' and `\#\#o'.\footnote{For a full discussion of the limits of tokenisation, see \citet{Mielke:21}.} We are therefore interested in alternative approaches which use lightweight tokenisation schemes (in particular character-level segmentation) coupled with powerful neural-network based composition functions to build language models (see Section \ref{sec:benefits} for a discussion of the benefits of character-level modelling). In this paper, we aim to determine the best way to build such models, focussing on models which output vector representations for each input character.

However, as the field of pretrained character-level modelling is relatively new, comparisons are complicated by the fact that recently proposed methods use different model architectures, pretrain on different data using different training objectives, and evaluate on different downstream tasks. With so many variables changing simultaneously, it is difficult to disentangle the effect of each individual choice in the modelling pipeline, and therefore decide on an overall best model configuration. To answer this question, we tested many model architectures and pretraining objectives from recent literature on a unified set of evaluation tasks, with the same training procedure. We identify one particular configuration that shows the best performance across all of our downstream evaluation tasks, namely a combination of the Charformer downsampling model \citep{Tay:2021}, and CANINE upsampling model and pretraining procedure \citep{Clark:22}. We dub this configuration \textbf{BORT}, for \textbf{B}idirectional \textbf{O}rthographic \textbf{R}epresentation \textbf{T}echnique. This model even outperforms a BERT baseline on all tasks we consider, while being moderately slower to pretrain (\S \ref{sec:big_comparison}).

One sticky point we discovered is that the best modelling configuration we found above relies crucially on a tokeniser during pretraining. We investigate alternative objectives that do not require a tokeniser, and find that these objectives result in worse-performing models. In addition, we also investigate the impact of the tokeniser used to pretrain the model, and find that the quality of the tokeniser (measured by vocabulary size) has a big impact on the final model downstream task performance, even though the tokeniser is not used at all during evaluation. This results in the unfortunate situation that users of such models have a hidden dependency on the tokeniser used to train the model; hence, users may be using models out of domain without any explicit feedback (such as worse tokeniser compression rates), causing difficult-to-detect performance regressions.

Taken together, we believe our results show that character-level representation models are ready to supplant subword-level models as a default choice for converting text into features. However, these models still require extensive supervision from tokenisers, and we believe that the next frontier of research in character-level modelling is finding ways to once and for all eliminate tokenisation as a key step in the NLP pipeline.

\section{The ingredients to make a character-level encoder model}

In this section, we aim to give an overview of all the components necessary to make a performant and efficient encoder-only model which operates on characters and outputs contextualised character representations. Working with characters rather than subword tokens brings many challenges, which have been solved in different ways in prior literature; we compare the selected methods in our experiments. In the following section, words in \textbf{bold} correspond to one cell in Table \ref{tab:my_label}.

\subsection{Input feature representation}
\label{sec:chars_or_bytes}

The first design choice that must be made when moving away from subword-based tokens is the input granularity. Typically, there are two choices: either (Unicode) \textbf{characters} \citep{Zhang:15,Kim_Jernite_Sontag_Rush_2016,ling-etal-2015-finding}, or the underlying \textbf{byte representation} \citep{gillick-etal-2016-multilingual}. The advantage of using bytes is the compact vocabulary (there are only 256 bytes); the disadvantage is that many Unicode characters require multiple bytes to encode, which further inflates the sequence length. Indeed, all non-ASCII characters require multiple bytes to encode in UTF-8. This disproportionately impacts non-European scripts, potentially harming the performance of multilingual byte-level models on such languages. In our current work, we exclusively use characters.

The downside of working with Unicode characters is the extremely large vocabulary: there are 1,114,112 code points allocated in 17 planes, each with 65,536 characters. \citet{Clark:22} solve the large vocabulary issue by using \textit{hash embeddings}, which compactly map the entire vocabulary to fixed-size vectors. However, as these embeddings are random, they cannot take advantage of representation learning at the orthographic level. Learnt character embeddings can help associate variations of the same character (e.g. \textit{a} and \textit{\"a}) and phonetically similar characters from different scripts (e.g. \textit{r} and \textit{$\rho$}). Further, the orthographic units of some scripts (e.g. Chinese characters) may themselves be semantically informative. We therefore add \textbf{learnt embeddings} for the Basic Multilingual Plane, which covers almost every script used to write modern languages.

\subsection{Architecture}

\begin{figure*}[t]
     \centering
     \begin{subfigure}[t]{0.3\linewidth}
         \centering
         \includegraphics[width=\textwidth]{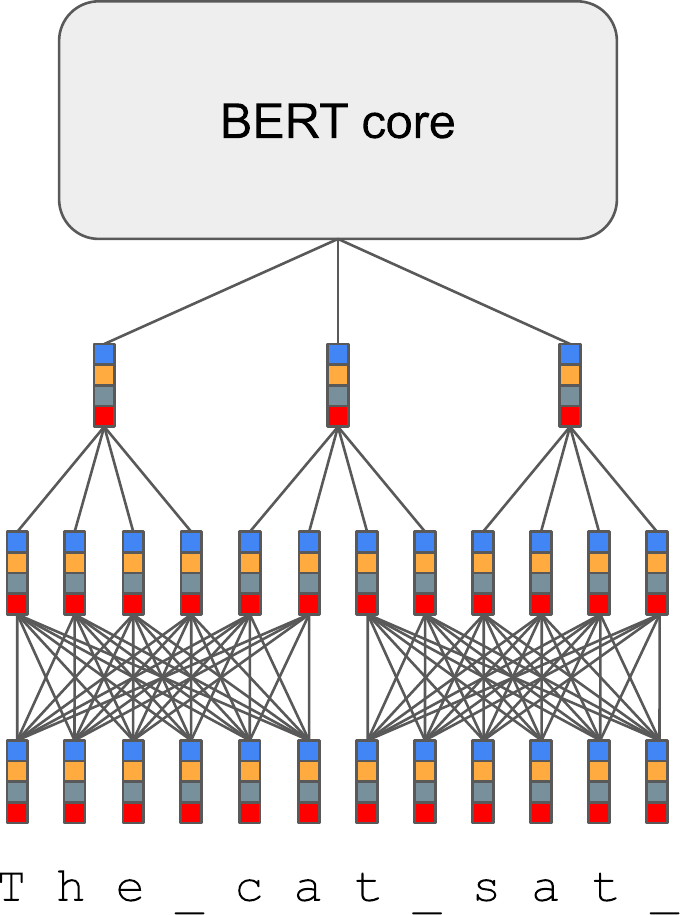}
         \caption{CANINE downsampling}
         \label{fig:canine_downsampler}
     \end{subfigure}
     \hfill
     \begin{subfigure}[t]{0.3\linewidth}
         \centering
         \includegraphics[width=\textwidth]{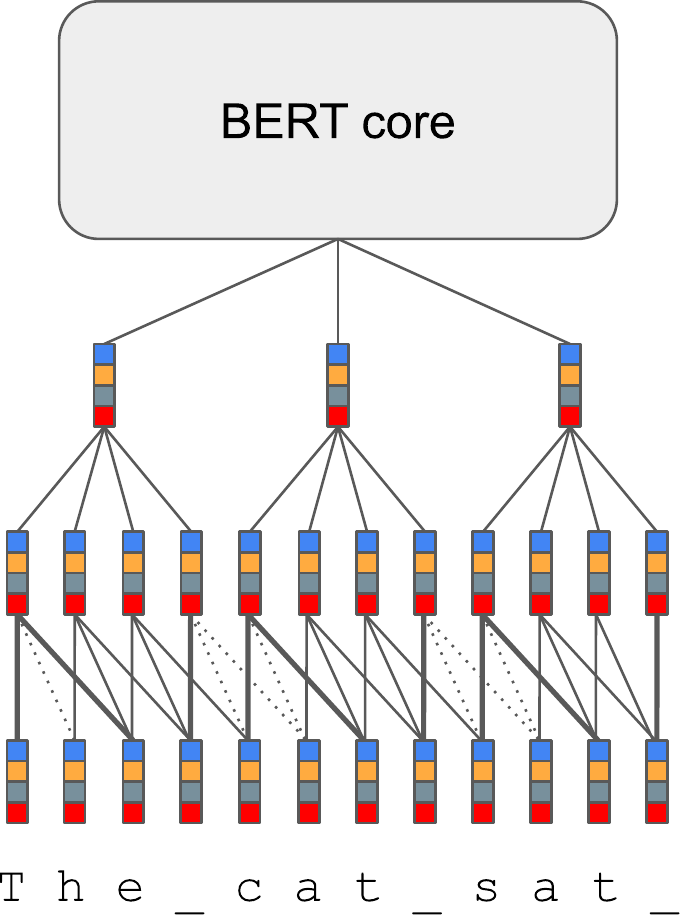}
         \caption{Charformer downsampling}
         \label{fig:charformer_downsampler}
     \end{subfigure}
     \hfill
     \begin{subfigure}[t]{0.3\linewidth}
         \centering
         \includegraphics[width=\textwidth]{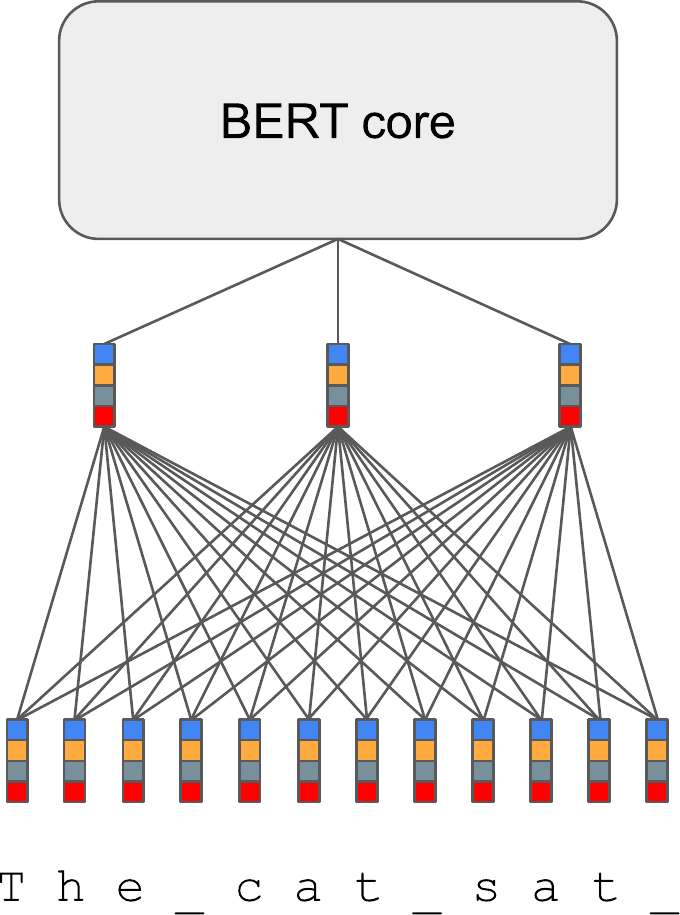}
         \caption{Perceiver downsampling}
         \label{fig:perceiver_downsampler}
     \end{subfigure}
        \caption{A visual comparison of the downsampling architectures we consider. Lines represent information flow between representations (either convolutions or restricted attention); the dashed lines in Fig. \ref{fig:charformer_downsampler} represent attention weights over different convolution widths.}
        \label{fig:downsampling_architectures}
\vspace{-1.5em}
\end{figure*}

\begin{figure}[t]
     \centering
     \begin{subfigure}[t]{0.48\linewidth}
         \centering
         \includegraphics[width=\textwidth]{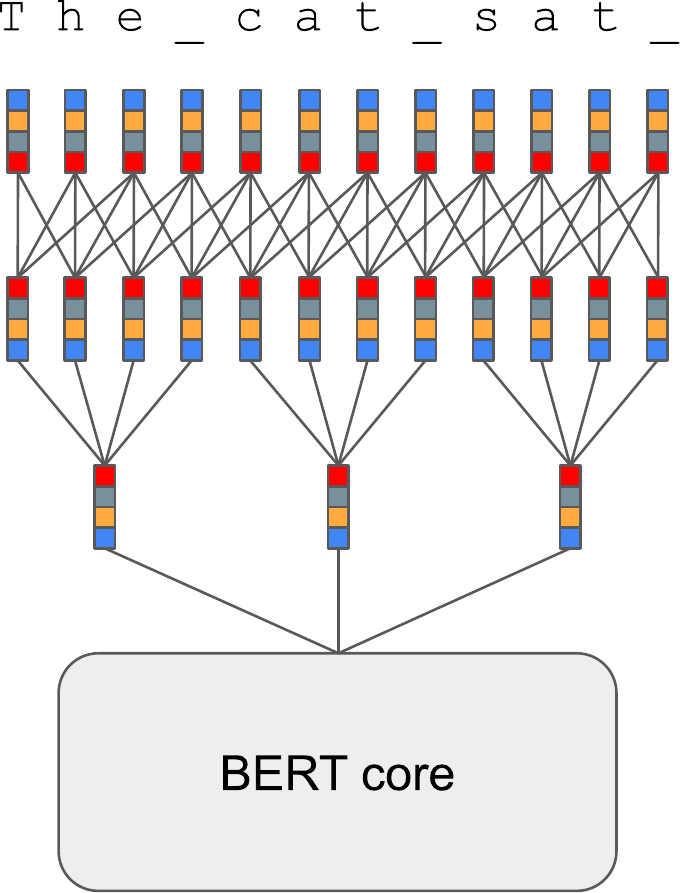}
         \caption{CANINE upsampling}
         \label{fig:canine_upsampler}
     \end{subfigure}
     \hfill
     \begin{subfigure}[t]{0.48\linewidth}
         \centering
         \includegraphics[width=\textwidth]{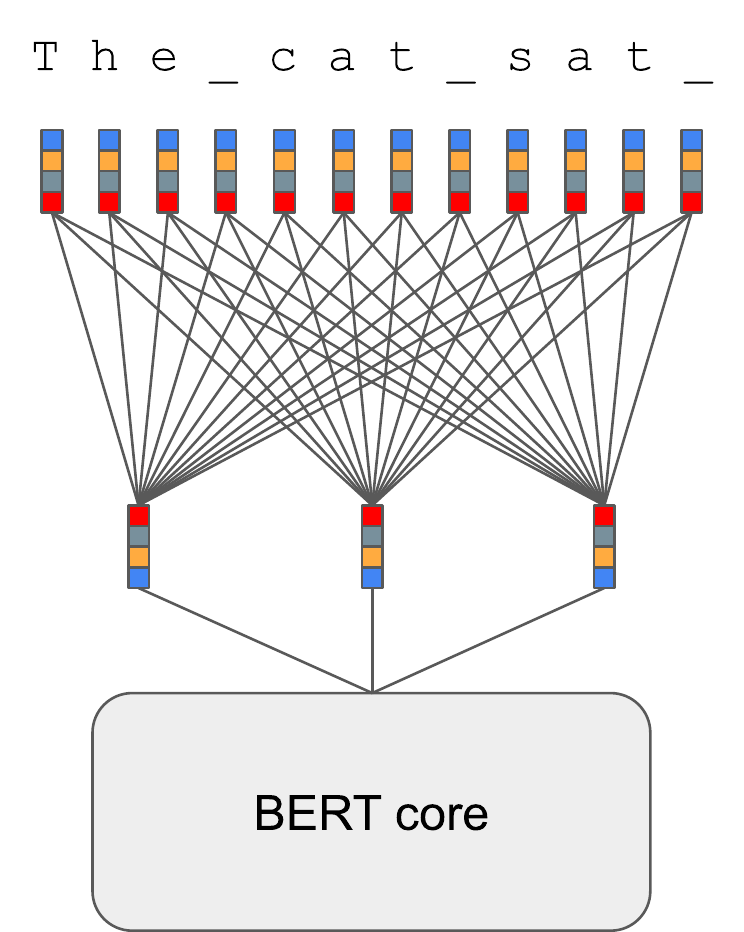}
         \caption{Perceiver upsampling}
         \label{fig:perceiver_upsampler}
     \end{subfigure}
    \caption{A visual comparison of the upsampling architectures we consider.}
    \label{fig:upsampling_architectures}
\vspace{-2em}
\end{figure}

One fundamental limitation of working directly with characters rather than subword tokens is that a longer sequence length is required to maintain the same context window. \citet{Clark:22} find that typically, a 4x larger sequence length is needed. However, as attention is quadratic in input length, it is not typically feasible to directly apply the standard deep Transformer model architecture to characters. Instead, the character sequence is usually first \textit{downsampled} into a more manageable length, and then processed, typically with a stack of Transformer layers similar to BERT \citep{devlin-etal-2019-bert}. The output of the BERT core is then \textit{upsampled} back to the original sequence length to obtain the final model outputs. We discuss both stages in more detail subsequently, and examine the overall performance and data efficiency of different model architectures in Section \ref{sec:big_comparison}.

\paragraph{Downsampling}
\label{para:downsampling}
The downsampling operation is often thought of as analogous to tokenisation in the standard NLP pipeline, as it combines multiple characters into a single representation in a similar way to how tokenisers segment text into multi-character model inputs. Many different downsampling architectures have been proposed--in this paper we examine three: \textbf{Perceiver} \citep{jaegle2022perceiver}, \textbf{CANINE} \citep{Clark:22} and \textbf{Charformer} \citep{Tay:2021}.

With these three models, it is further possible to split the downsampling stage into two separate steps: a \textit{contextualisation} stage which aggregates information across multiple characters, and a \textit{pooling} stage that compresses the character sequence. CANINE first uses a windowed local attention over the input character sequence to aggregate information among neighbouring characters, before using a strided 1D convolution with filter width 4 and stride 4 to achieve the 4x downsampling rate. By contrast, Charformer first applies an attention-weighted sum over convolutions of varying widths at each position, before compressing the contextualised characters using average pooling, again using window size 4 and stride 4. Perceiver is the exception as it has no separate contextualisation stage; instead, it directly downsamples the embedded character sequence with a cross-attention layer, using a learnt bank of latent query vectors. We illustrate these architectures in Figure \ref{fig:downsampling_architectures}. 

\paragraph{Upsampling}
\label{para:upsampling}

Conceptually, a simple method to go from token embeddings to character embeddings is to repeat each contextualised token embedding $N$ times, where $N$ is the length (in characters) of the token. With such embeddings, it is trivial to match the performance of the token-level model by projecting the token-level span to the character-level span. Indeed, the \textbf{CANINE upsampling} layer repeats each output of the downsampled BERT core 4 times (to match the downsampling rate), concatenates the repeated latent representations with the contextualised character embeddings, applies a convolution over these, and then applies a final all-to-all Transformer layer. By contrast, \textbf{Perceiver} applies a cross-attention operation between the output of the deep Transformer stack and a bank of query vectors the same length as the original character sequence. Both architectures are illustrated in Figure \ref{fig:upsampling_architectures}.\footnote{\citet{Tay:2021} introduce Charformer in an encoder-decoder framework with a token-level decoder, so there is no Charformer upsampler.}

\subsection{Pretraining objectives}
\label{sec:pretraining_objectives}
The typical pretraining objective for language representation models is masked language modelling -- given some input text, the model must learn to reconstruct masked-out portions given the context. For subword-level models, the masked portion is often a single token, although alternative masking schemes exist \citep{joshi-etal-2020-spanbert,levine2021pmimasking}. However, masking individual characters does not give a good pretraining objective, as individual characters are very easy to predict given their surrounding context. We therefore investigate alternative masking schemes, and prediction targets derived from such masking schemes, and we outline the ones we consider in this section.

\paragraph{Masking schemes}
\label{para:masking_schemes}
As masking individual characters does not train good models, most masking schemes pick spans of characters to mask instead. The simplest method is to mask \textbf{random} spans of characters \citep{xue-etal-2022-byt5,Keren:22}. However, \citet{levine2021pmimasking} showed that the best spans to mask are those with a high coherence, which random spans do not have. A better masking scheme is to mask semantically meaningful spans. One heuristic to pick such spans is to use \textbf{whitespace} \citep{jaegle2022perceiver}; unfortunately, many orthographies around the world do not use whitespace, which reduces the cross-linguistic portability of this scheme. Another heuristic is to use a \textbf{tokeniser} to decide which character spans to mask, but predict the masked characters instead. This method is language independent, but has the downside that it reintroduces a dependency on an external tokeniser, which was a motivation to move to character-level modelling in the first place.

\paragraph{Prediction targets}
\label{para:prediction_targets}

Once a span of characters has been masked, one must derive a prediction target from the masked span. If a tokeniser-based masking scheme is used, one can simply predict the masked token using a classifier head. This is the CANINE-S training scheme from \citet{Clark:22}. However, if the random or whitespace masking schemes are used, the set of possible masked spans is too large to classify directly. In this case, we can back off to predicting the characters of the masked span. This can either be done \textbf{autoregressively} (with predicted characters being revealed one-by-one) as in CANINE-C, or \textbf{independently }(with each character prediction being made without knowledge of the other masked characters; \citealt{jaegle2022perceiver,Keren:22}). Predicting characters has the additional complication that the Unicode vocabulary is very large. We therefore use the same hashing trick that we use to compactly represent Unicode characters: we hash the Unicode codepoint of a character 8 ways at random, and then predict each hash independently.

\section{Evaluation}

\subsection{Evaluation tasks}

Previous works in the space of character-level representation learning have all chosen distinct evaluation tasks, which makes direct comparison across different methods difficult. We compare all our models on the same evaluation tasks, which we split into two groups: probing tasks and downstream tasks. For the probing tasks, we fix the model parameters and learn a classifier to predict morphological features and part-of-speech tags, which we take from Universal Dependencies \citep{nivre2020universal}. We use information-theoretic probing \citep{voita-titov-2020-information} to assess how easily extractable morphological information is from each model---specifically, we use the prequential codelength probing procedure. We are interested in whether character-based models represent morphological information in a more easily extractable way than subword-based models---one perceived benefit of character-level models is that they may be able to represent morphology better \citep{vania-lopez-2017-characters}, which could lead to better performance on morphologically rich languages.

The second group of tasks are downstream tasks more aligned with typical NLP model use cases. We use WikiANN NER \citep{pan-etal-2017-cross} and extractive QA (TyDi-QA gold passage; \citealt{clark-etal-2020-tydi}) to represent both sequence labelling and span extraction tasks which require information to be localised at specific locations in the text. Character-level models have previously shown to perform well at general sentence representation tasks, such as GLUE \citep{jaegle2022perceiver,Clark:22}; however, CANINE performed poorly at high-resource NER in particular, and so our choice of WikiANN on our evaluation languages set a high bar for the character-level models. We believe that tasks like QA and NER require more higher-level semantically oriented information, and we would like to demonstrate that it is possible to learn such information directly from characters.


We evaluate gold passage TyDI-QA in the standard way (macro-averaged F1 across languages excluding English). For UD probing and WikiANN NER, we evaluate on a typologically diverse choice of languages: Arabic, English, Finnish, German, Hungarian, Indonesian, Italian, Russian and Turkish, and report metrics macro-averaged across all languages, including English.

\section{Experiments}
We train all our models on the same multilingual Wikipedia dump as \textsc{mBERT}, with the same exponentially weighted language sampling strategy. Our baseline model architecture is BERT-base with 110M parameters; all other models are comparable in size. We train each model using 32 TPUv3 chips for 250k  steps with total batch size 3072. Models took between 3 and 4 days to complete training. We found the batch size parameter crucial for final model performance: using a smaller batch size degraded final model performance, while character-level model performance was unstable at a larger batch size. For exact pretraining hyperparameters and downstream task evaluation procedures, please see Appendices \ref{sec:hyperparams} and \ref{sec:evaluation}. Unless otherwise stated, the hyperparameters are constant across all experiments; each experiment aims to examine the influence of a specific choice of variable. We evaluate model checkpoints on a rolling basis during training on all our evaluation tasks, and select the model checkpoint which performs the best on TyDi-QA.

\subsection{Model architecture comparison}
\label{sec:big_comparison}
\begin{table*}[t]
    \centering
    \begin{tabular}{l l l c c c c}
        \toprule
         & & &  \multicolumn{2}{c}{Probing tasks} & \multicolumn{2}{c}{Downstream tasks} \\
         & & Downsampler & UD Feats. $\downarrow$ & UD POS $\downarrow$ & TyDi-QA F1 $\uparrow$ & WikiANN F1 $\uparrow$ \\
        \cmidrule(lr){2-7}
        \multirow{6}{*}{\rotatebox[origin=c]{90}{Upsampler}} & \multirow{3}{*}{\rotatebox[origin=c]{90}{CANINE}} & CANINE & 2.55 $\pm$ 0.00 & 1.35 $\pm$ 0.02 & 76.09 $\pm$ 0.47 & 89.10 $\pm$ 0.18 \\
         & & Charformer & 2.72 $\pm$ 0.03 & 1.49 $\pm$ 0.02 & \textbf{78.76 $\pm$ 0.56} & \textbf{90.65 $\pm$ 0.02} \\
         & & Perceiver & 2.53 $\pm$ 0.00 & 1.34 $\pm$ 0.02 & 75.51 $\pm$ 0.42 & 89.79 $\pm$ 0.07 \\
        \cmidrule(lr){3-7}
         & \multirow{3}{*}{\rotatebox[origin=c]{90}{Perceiver}} & CANINE & 2.47 $\pm$ 0.00 & 1.33 $\pm$ 0.01 & 68.00 $\pm$ 1.26 & 88.16 $\pm$ 0.04 \\
         & & Charformer & 2.49 $\pm$ 0.01 & 1.39 $\pm$ 0.01 & 69.52 $\pm$ 0.45 & 82.50 $\pm$ 0.29 \\
         & & Perceiver & \textbf{2.44 $\pm$ 0.01} & \textbf{1.30 $\pm$ 0.00} & 73.17 $\pm$ 0.41 & 89.66 $\pm$ 0.01 \\
         \midrule
         \multicolumn{3}{c}{BERT Baseline} & 2.63 $\pm$ 0.01 & 1.35 $\pm$ 0.00 & 76.97 $\pm$ 0.90 & 90.29 $\pm$ 0.05 \\
        \bottomrule
    \end{tabular}
    \caption{Comparison between different character-level modelling architectures on our chosen evaluation task suite. All results are macroaveraged across languages. UD feature and part-of-speech probing is measured in nats/label (lower is better). TyDi QA and WikiANN NER performance is reported in F1 (higher is better). All results are averaged over 3 finetuning runs with different random seeds.}
    \vspace{-1em}
    \label{tab:big_comparison}
\end{table*}

\begin{table}[t]
    \centering
    \resizebox{\columnwidth}{!}{
    \begin{tabular}{l l l c c}
        \toprule
         & & Downsampler & Throughput & FLOPS \\
        \cmidrule(lr){2-5}
        \multirow{6}{*}{\rotatebox[origin=c]{90}{Upsampler}} & \multirow{3}{*}{\rotatebox[origin=c]{90}{CANINE}} & CANINE & 0.68x & 2.01x \\
         & & Charformer & 0.68x & 2.70x \\
         & & Perceiver & 0.81x & 1.91x \\
        \cmidrule(lr){3-5}
         & \multirow{3}{*}{\rotatebox[origin=c]{90}{Perceiver}} & CANINE & 0.72x & 1.51x \\
         & & Charformer & 0.72x & 2.21x \\
         & & Perceiver & 0.85x & 1.39x \\
         \midrule
         \multicolumn{3}{c}{BERT} & 1x & 1x \\
        \bottomrule
    \end{tabular}
    }
    \caption{Computational efficiency of the models we consider, relative to token-level BERT. Throughput refers to pretraining examples per second, while FLOPs is of a forward model pass on a single example.}
    \vspace{-1em}
    \label{tab:compute_comparison}
\end{table}

\begin{table}[t]
    \centering
    \resizebox{\columnwidth}{!}{
    \begin{tabular}{l l l c c}
        \toprule
         & & Downsampler & TyDi-QA & WikiANN \\
        \cmidrule(lr){2-5}
        \multirow{6}{*}{\rotatebox[origin=c]{90}{Upsampler}} & \multirow{3}{*}{\rotatebox[origin=c]{90}{CANINE}} & CANINE & 72.00 & 88.04 \\
         & & Charformer & \textbf{75.56} & \textbf{89.78} \\
         & & Perceiver & 67.44 & 86.21 \\
        \cmidrule(lr){3-5}
         & \multirow{3}{*}{\rotatebox[origin=c]{90}{Perceiver}} & CANINE & 64.32 & 86.27 \\
         & & Charformer & 66.90 & 80.85 \\
         & & Perceiver & 68.34 & 85.12 \\
         \midrule
         \multicolumn{3}{c}{BERT} & 73.82 & 89.06 \\
        \bottomrule
    \end{tabular}
    }
    \caption{Comparison of the data efficiency of the models we consider. All numbers are normalised area-under-F1 curves during model training on the respective task.}
    \vspace{-1em}
    \label{tab:model_data_efficiency}
\end{table}

We first report a cross-model comparison between BERT, CANINE and Perceiver on our set of evaluation tasks. For these comparisons, we use the tokeniser-based masking scheme with the mBERT WordPiece tokeniser, and predict the masked tokens from a closed vocabulary. Our results are shown in Table \ref{tab:big_comparison}.

\paragraph{Character-level models do better at morphology (usually)}
Our results show that most of the character-level models outperform BERT on the morphological probing tasks. This result is in line with existing literature on the benefits of character-level features for low-level NLP tasks \citep{vania-etal-2018-character}. The only exception is the Charformer-CANINE model combination, which however does well on the more downstream tasks. We discuss this more in the next section.

\paragraph{Charformer-CANINE surpasses BERT at downstream tasks}
On our downstream semantically-oriented evaluation tasks (TyDI-QA and WikiANN NER), we note that the combination of Charformer encoder and CANINE decoder outperforms our retraining of the BERT baseline model on both QA and NER, without using additional features such as character n-grams. We believe this result shows that with the right architecture and training objective, current-generation character-level models exceed the performance of token-based models and should be considered as a new default choice for extracting contextual embeddings from text.

One interesting aspect of our results is that model performance on the UD morphological feature tagging probe task tends to be anti-correlated with performance on the downstream tasks. Indeed, the Spearman correlation across all models between UD Feats and TyDi-QA F1 is 0.89 and between UD Feats and WikiANN F1 is 0.68. One explanation for this might be that as models learn to compose characters into more `semantic' units, less information about individual characters is propagated through the model, and that there is a trade-off between representing low-level morphological information vs higher-level semantic information. Indeed, there is evidence that character-level models tend to oversmooth based on orthographic similarity \citep{cao-rei-2016-joint,Josefowicz:16}, and character n-gram features have been used to try and circumvent this \citep{bojanowski-etal-2017-enriching,Clark:22}. Charformer-CANINE is able to perform well without such n-gram features, and this may be that the convolutions over characters implicitly represent character n-grams well already.

\paragraph{Character-level models are less compute efficient}
\label{sec:compute_efficiency_results}

We next evaluate the compute efficiency of our different model architectures. We compare two main quantities: pretraining throughput (in examples/sec) and FLOPs per forward pass on a single example. In general, more FLOPs is associated with better model performance \citep{Kaplan:20,hoffmann2022an} at the cost of inference speed, but due to hardware design, not all FLOPs are created equal. We show the results in Table \ref{tab:compute_comparison}. As all our character-level models are built around the BERT core, it is expected that every model compares unfavourably to BERT on these metrics. We note that even though the Charformer-CANINE model (which performs the best overall) uses the most FLOPs per forward pass, its pretraining throughput is not proportionally slower, suggesting that the model architecture is efficient to run on current-generation hardware.

\paragraph{Model architecture impacts data efficiency}

To perform model selection based on downstream task performance, we evaluate these tasks over the course of model pretraining. This lets us probe how data-efficient each model is during pretraining, which can give us indications about whether the intrinsic biases of the model are suited to learning general linguistic information.

We evaluate using area-under-training-curve metrics, similar to prequential coding \citep{Blier:18,Yogatama:19,voita-titov-2020-information}. Prequential coding can be viewed as area under the log-loss training curve; we instead measure area under the F1 curve, normalised by the total number of training steps. We present our results in Table \ref{tab:model_data_efficiency}. From these numbers, one can see that the lack of innate bias in the Perceiver model components renders it less data efficient. We note that a core feature of theories of linguistic morphology is that morphemes consist of units close together \citep{Haspelmath2010-nz}; the authors are unaware of any theory of morphology that allows arbitrary long-range word formation. The Perceiver downsampling mechanism on the other hand can potentially aggregate information from any character combination into a single unit, and hence it has to learn a preference to compose nearby characters, rendering it less data-efficient. By contrast, both CANINE and Charformer inherently combine adjacent characters together to form latent representations. Indeed, the difference between the numbers in Table \ref{tab:big_comparison} and Table \ref{tab:model_data_efficiency} for the Perceiver-CANINE model is particularly great, and one can see an obvious `kink' in the training curve for this model as it discovers the necessary biases for combining characters into higher level units.

\paragraph{Learnt character embeddings improve results}
If we remove the learnt character embeddings and rely solely on hash embeddings, results for TyDi-QA drop to 64.48 $\pm$ 24.56, and for WikiANN drop to 87.98 $\pm$ 0.06. The large variance in TyDi results is caused by one finetuning run achieving a very low F1. This shows that learnt character embeddings not only result in better overall task performance, but also result in more stable models. Character embeddings have been shown to capture information such as phonetics and shape \citep{boldsen-etal-2022-interpreting}, which can be helpful features for the model. We therefore recommend using learnt character embeddings in all character-level models. 

\subsection{Masking scheme and pretraining objective}
\label{sec:loss_fn_results}
\begin{table}[t]
    \centering
    \resizebox{0.48\textwidth}{!}{
    \begin{tabular}{l l l c c}
        \toprule
         & & Masking & TyDi-QA & WikiANN \\
        \cmidrule(lr){2-5}
        \multirow{6}{*}{\rotatebox[origin=c]{90}{Prediction targets}} & \multirow{3}{*}{\rotatebox[origin=c]{90}{Auto.}} & Random & 75.20 $\pm$ 0.80 & 86.70 $\pm$ 0.57 \\
         & & Tokeniser & 76.46 $\pm$ 1.19 & 89.64 $\pm$ 0.26 \\
         & & Whitespace & 77.66 $\pm$ 0.71 & 88.68 $\pm$ 0.33 \\
        \cmidrule(lr){3-5}
         & \multirow{3}{*}{\rotatebox[origin=c]{90}{Indep.}} & Random & 72.76 $\pm$ 0.17 & 87.48 $\pm$ 0.12 \\
         & & Tokeniser & 73.67 $\pm$ 0.55 & 88.35 $\pm$ 0.17 \\
         & & Whitespace & \textbf{78.92 $\pm$ 0.19} & \textbf{89.95 $\pm$ 0.02} \\
        \bottomrule
    \end{tabular}
    }
    \caption{Comparison of alternative token-prediction-free pretraining objectives given by combining a method of selecting spans of characters to mask and how to predict the masked characters.}
    \vspace{-1em}
    \label{tab:loss_fn_comparison}
\end{table}

\begin{table}[t]
    \centering
    \resizebox{0.48\textwidth}{!}{
    \begin{tabular}{l l l c c}
        \toprule
         & & Masking & TyDi-QA & WikiANN \\
        \cmidrule(lr){2-5}
        \multirow{6}{*}{\rotatebox[origin=c]{90}{Prediction targets}} & \multirow{3}{*}{\rotatebox[origin=c]{90}{Auto.}} & Random & 71.93 & 85.66 \\
         & & Tokeniser & 73.88 & \textbf{88.83} \\
         & & Whitespace & 74.39 & 88.10 \\
        \cmidrule(lr){3-5}
         & \multirow{3}{*}{\rotatebox[origin=c]{90}{Indep.}} & Random & 65.48 & 77.19 \\
         & & Tokeniser & 67.61 & 80.07 \\
         & & Whitespace & \textbf{74.46} & 88.26 \\
        \bottomrule
    \end{tabular}
    }
    \caption{Comparison of the data efficiency of alternative pretraining objectives. All numbers are normalised area-under-F1-curves during model training.}
    \vspace{-1em}
    \label{tab:loss_fn_data_efficiency_comparison}
\end{table}

In this section, we investigate whether it is possible to use the tokeniser-free masking schemes and prediction targets introduced in Section \ref{sec:pretraining_objectives} to train models which perform as well as tokeniser-based models. We focus here on the Charformer-CANINE model which showed promise in the previous section, and train it in the same setting, using each combination of masking scheme and character-level prediction target. We show the results in Table \ref{tab:loss_fn_comparison}. As one can see, no combination of masking scheme and prediction targets uniformly surpass the performance of the tokeniser-based model. Indeed, the performance disparity is particularly stark on WikiANN NER, which is a task requiring heavy memorisation, suggesting the bias of predicting discrete tokens helps the model discover units of language amenable to memorisation. 

It is still possible to observe consistent internal variation between the different masking schemes. Random masking performs the worst of the masking schemes, suggesting that it is important to mask semantically coherent spans of characters. Further, whitespace masking performs better than tokeniser-assisted masking, giving more evidence that tokenisation with a fixed vocabulary bottlenecks language model training. Finally, it appears that in general autoregressive character prediction performs better than independent character prediction when a suboptimal masking scheme is used.

We also examine the data efficiency of character-level prediction targets. Table \ref{tab:loss_fn_data_efficiency_comparison} shows that autoregressive prediction is a lot more stable during model training than independent character prediction for suboptimal masking schemes. Further, comparing the numbers in Table \ref{tab:loss_fn_data_efficiency_comparison} to Table \ref{tab:model_data_efficiency} shows that training models using token-level predictions is more data efficient, and suggests that token-level targets are better suited to learning linguistic information. We therefore believe that more work is necessary to discover better ways to predict open-vocabulary masked targets that combine the flexibility of character-level prediction and the intrinsic bias of fixed morpheme-like units.

\subsection{Tokeniser quality}
\label{sec:tokeniser_sweep}
\begin{table}[t]
\resizebox{\columnwidth}{!}{
    \centering
    \begin{tabular}{l l r r r r r}
        \toprule
         & & \multicolumn{4}{c}{Vocabulary size} \\
         & Model & 10,000 & 25,000 & 50,000 & 100,000 \\
        \cmidrule(lr){2-6}
        \multirow{2}{*}{\rotatebox[origin=c]{90}{QA}} & Subword & 68.24 & 74.20 & 73.97 & 76.68 \\
         & Character & 66.38 & 70.23 & 76.93 & 79.11 \\
        \\
         \multirow{2}{*}{\rotatebox[origin=c]{90}{NER}} & Subword & 89.65 & 90.02 & 90.21 & 90.34 \\
         & Character & 88.01 & 89.66 & 90.37 & 90.95 \\
        \bottomrule
    \end{tabular}
    }
    \caption{The effect of varying tokeniser size on downstream task performance. Subword refers to a model trained with subword inputs, which character refers to a character-input model. All results are F1 scores.}
    \vspace{-1em}
    \label{tab:tokeniser_size}
\end{table}
Finally, since we showed that using a tokeniser still gives the best results when pretraining character-level models, it is natural to ask how much the quality of the tokeniser influences the resulting model. We train SentencePiece unigram tokenisers \citep{kudo-richardson-2018-sentencepiece} of varying vocabulary sizes (as a proxy of tokeniser quality) on a subset of the pretraining data. We then train BERT and Charformer-CANINE models using these tokenisers, and provide the results in Table \ref{tab:tokeniser_size}.

Larger vocabulary sizes consistently lead to better downstream task performance for both models, even the character-level model. This result is even more remarkable given that the tokeniser is only used for pretraining and discarded on downstream fine-tuning. Therefore, users of character-level models have a hidden long-distance dependency on the tokeniser that was used to train the model, even though this is not exposed to the user. We feel this state of affairs is extremely unfortunate, as a substandard pretraining-time tokenisation can have a large impact on downstream performance yet be completely invisible to the user.

Further, we note that we do not appear to have reached the limit of model improvement due to increasing the vocabulary size. The maximum size we considered is 100,000, due to resource constraints, but we note that larger vocabularies have been considered in multilingual representation learning (\citet{conneau-etal-2020-unsupervised} use a vocabulary size of 250,000, for instance). We believe that efficient ways of scaling up vocabulary size even further is an interesting avenue of research.

\section{Discussion}
\subsection{Benefits of character-level modelling}
\label{sec:benefits}

We have shown that character-level models can achieve better performance at a range of tasks than token-level models, at the cost of slightly slower models. We believe this tradeoff is worth making, and we outline the advantages of character-level modelling in this section.

\paragraph{Removing tokenisers from the NLP pipeline}
We believe that tokenisation imparts a significant engineering burden on users of NLP models. Tokenisers are themselves parametric models, and different tokeniser settings can have a large impact on task performance \citep{bostrom-durrett-2020-byte}. Further, there is evidence that language model performance is bottlenecked by tokeniser suboptimality due to e.g. poor out-of-domain performance \citep{cao-rimell-2021-evaluate}. In addition, tokenisation can introduce hidden bugs due to differences in capitalisation, whitespace or other special characters. For all of these reasons, we believe that removing tokenisation from NLP pipelines improves the experience of using language models.

\paragraph{Annotation is easier at the character level}
As characters are the natural unit of orthography, it is typically easier to annotate tasks, especially span-extraction tasks, at the character level. This is especially true for scripts which do not use whitespace in their orthography, or when whitespace and syntactic tokens do not match. Indeed, gold passage TyDi-QA drops data from Thai and Japanese so that the standard \texttt{run\_squad.py} script can be used. These implicit data selection effects can systematically bias experimental results---for instance, we believe that whitespace masking would work less well on non-whitespace languages, yet none are in the set of languages we evaluate on. We therefore believe that annotating tasks at the token-level for modelling convenience is a mistake, and we believe that annotation should be performed with linguistic validity as the main motivation.

\subsection{Inductive bias, model architecture and training procedure}
How low-level linguistic units combine into meaningful higher-level units is one of the best-studied areas of linguistics, and we know many of the basic cross-lingual rules of building morphemes. It is therefore interesting that the model architecture and training procedure which worked the best are also those which conform most to existing knowledge about morphology. The Charformer encoder and CANINE decoder both make strong locality assumptions about how characters combine, and the Charformer encoder explicitly operates over segmentations of the input. In addition, the tokeniser-assisted training objective encodes information about units of language into the model. We believe our results show the importance of domain knowledge when building models, especially when compute or data efficiency is a requirement.

\section{Conclusion}
In this paper, we examined how best to train a character-level encoder model, and identified a recipe that produces models exceeding the performance of token-based models at a comparable compute cost, suggesting that the time of general purpose character-level modelling has arrived. 

\section*{Limitations}

\subsection*{Choice of languages}
Our choice of languages for WikiANN and UD probing evaluations were intended to strike a balance between being being typologically diverse and having data in our chosen benchmarks. However, there are major language families and geographical regions not represented in our languages (there is no indigenous language of the Americas in any of our benchmarks, and no southern African language in UD or WikiANN). While we expect the trends in our results to continue to hold for other languages, we believe that further investigation is necessary on more languages to confirm our hypothesis.

\subsection*{Choice of evaluation tasks}
One notable omission from our evaluation suite are sentence-level tasks, such XNLI \citep{conneau-etal-2018-xnli}, XGLUE \citep{liang-etal-2020-xglue} and cross-lingual retrieval tasks. One reason is that previous work has shown that character-level models already perform well on these evaluations. In our work, we were particularly interested in situations where prior work showed character-level models underperforming subword-based models. In particular, CANINE underperformed at NER, especially in the high-resource CoNLL 2003 NER dataset \citep{tjong-kim-sang-de-meulder-2003-introduction}. Therefore, we chose to focus specifically on NER and extractive QA as typical use cases of encoder-only models. In future work, we will investigate more thoroughly the capabilities of character-level models on a wider range of tasks.

\section*{Ethics statement}
Our work compares existing work on character-level language modelling, and we do not anticipate that it introduces any new risks beyond those introduced by the work we build on.

\section*{Acknowledgements}
We would like to thank Laura Rimell and Dan Garrette for extensive comments and advice throughout the duration of this project, as well as Valentin Hofmann and Paul Michel for comments on earlier versions of this paper. We would also like to thank the DeepMind language team for helpful discussions.

\bibliography{anthology,custom}

\appendix

\section{Training hyperparameters}
\label{sec:hyperparams}

\subsection{Model architectures}
Our standard model architecture is BERT-small. This uses 12 Transformer layers with hidden size 768, and 12 self-attention heads per layer. We use a context sequence length of 512 subword tokens for pretraining.

For the CANINE model, we use the same architecture as \citet{Clark:22}. We use a sequence length of 2048 characters during pretraining. The model consists of a local Transformer layer with context width 128 (i.e. each 128-width window of characters is processed independently) with hidden size 768 and 12 heads, followed by a strided convolution with width 4, stride 4 and output size 768 with a GeLU activation and layer normalisation. This results in a downsampled representation of length 512, which is then fed into a BERT-small core. For upsampling, we repeat the output of the inner Transformer 4 times (to match the downsampling rate) and concatenate this with the contextualised characters from downsampling model. We then run another convolution with filter width 4, stride 1 and output size 768, again followed by a GeLU activation and layer normalisation. Finally, we do an all-to-all Transformer layer to obtain the final output representation.

For Perceiver, we again use a sequence length of 2048 for pretraining. For the downsampling layer, we use an array of 512 randomly initialised vectors as the latent queries, and perform cross-attention using these query vectors and the character embeddings as the keys. The resulting downsampled representation of length 512 is then fed into a BERT-small-sized core, (which differs from the internal processing model of \citealt{jaegle2022perceiver}). To upsample, we used the contextualised character embeddings from the downsampling model as the query vectors to perform cross-attention with the output of the BERT core. We found that adding a skip connection between the character input and output helped the model learn more stably.

For Charformer, we used convolution filter widths in the range $[1, 2, 3, 4, 5]$. Rather than striding the convolution by the filter width, we densely applied the convolution (i.e. with stride 1), and do not apply the first 1D convolution. We computed attention weights for each convolution output at each character position with a 2 layer MLP with GeLU nonlinearity, and combined the output of the convolutions with these weights.

We also note that the placement of layer normalisation in attention layers for our model architectures was crucial for model performance \citep{pmlr-v119-xiong20b}. For self-attention, we found that post-norm worked the best, whereas for cross-attention pre-norm worked better. This mainly affected the Perceiver model, which uses cross-attention in the down- and up-sampling layers.

\subsection{Model implementation}
All models were implemented using JAX \citep{jax2018github} and Haiku \citep{haiku2020github}. We use a dropout rate of 0.1 after all matrix multiplications in the model. We use the LAMB optimizer \citep{You2020Large}, with a maximum learning rate of \num{1.25e-3}. We warm up the learning rate over the first 3125 training steps, and use a cosine decay learning rate scheduler \citep{Loshchilov:16} with length equal to the number of training steps and a final learning rate of \num{1.25e-5}. For our BERT baseline, we use a maximum learning rate of \num{1.8e-3} and a minimum of \num{1.8e-5}. We clip gradients to a maximum global norm of 10.0. We keep an exponential moving average of model weights during training with EMA parameter 0.9, updated after every 100 training steps, and evaluate using the average parameters. We found that this stabilised model training for the character-level models, and resulted in better task performance.

\section{Evaluation protocols}
\label{sec:evaluation}

\subsection{Probing tasks}
We use the prequential codelength probing paradigm of \citet{voita-titov-2020-information}, but follow a slightly different protocol. We use the training data of the largest UD dataset for each language we consider, and take a sample of 4000 sentences (or use the whole corpus if it is smaller then this), and split this into 10 shards. We then initialise a label prediction head and freeze the base model. We then sequentially evaluate each shard, before adding the shard to the training data for the tagging model. We then train the tagging model on batches of data randomly sampled from all shards that we have previously evaluated, and periodically evaluate on a dev set of data we set aside from the first shard. If the dev set loss stops improving, we then stop training and evaluate on the next shard, continuing in this way until we have evaluated every shard. We then add up the loss for all the evaluated shards and divide by the number of predictions to get the average codelength per tag. We use 2 V100 GPUs for training, and use a total batch size of 32. 

One difficulty with UD tagging tasks is that the tags are defined on syntactic tokens, which may not correspond to the surface form (for example, \textit{can't} is annotated as two syntactic tokens: \textit{can} and \textit{not}), and aligning syntactic tokens with the surface form may not be trivial. Further, tokenisation means that alignments between surface form tokens and the input to the model may also be non-trivial. However, most subword tokenisation schemes treat whitespace specially, and never merge tokens across whitespace. This means we can merge the UD morph and POS annotations for each syntactic token making up a whitespace token (i.e.~we merge the POS tags for \textit{can} and \textit{not} and tag \textit{can't} with this composite label), and predict this composite label as an atomic unit, at the cost of expanding the tagset. For all our tagging tasks, we take the first model token (either subword or character) corresponding to a whitespace token as the token representation and predict the tag based on the embedding of this token. Morphological features in UD are annotated as an unordered set of key-value pairs; we ignore this internal structure and treat each occurring set of tags as an atomic label. 

\subsection{TyDi-QA and WikiANN}
For these tasks, we finetune the full model. For both tasks, we use a single linear layer to produce the model logits over either BIO tags or start/end span indices. We combine the training data for all languages we consider, and train for 10 epochs for both tasks. We use 4 TPUv3 chips to finetune the model, and use a total batch size of 128. For TyDi-QA, we modify the official \texttt{run\_squad.py} script to accept non-WordPiece tokenisers (such as SentencePiece and character tokenisers). 




\end{document}